\documentclass[graybox, envcountchap]{svmult}

\usepackage{mathptmx}        
\usepackage{helvet}          
\usepackage{courier}         

\usepackage{makeidx}         
\usepackage{graphicx}        
\usepackage{multicol}        
\usepackage[bottom]{footmisc}

\makeindex             

\begin{document}

\frontmatter

%
%
%

\begin{dedication}
Use the template \emph{dedic.tex} together with the Springer document class SVMono for monograph-type books or SVMult for contributed volumes to style a quotation or a dedication\index{dedication} at the very beginning of your book in the Springer layout
\end{dedication}

%
%

\foreword

Use the template \textit{foreword.tex} together with the Springer document class SVMono (monograph-type books) or SVMult (edited books) to style your foreword\index{foreword} in the Springer layout. 

The foreword covers introductory remarks preceding the text of a book that are written by a \textit{person other than the author or editor} of the book. If applicable, the foreword precedes the preface which is written by the author or editor of the book.

\vspace{\baselineskip}
\begin{flushright}\noindent
Place, month year\hfill {\it Firstname  Surname}\\
\end{flushright}

%
%

\preface

Use the template \emph{preface.tex} together with the Springer document class SVMono (monograph-type books) or SVMult (edited books) to style your preface in the Springer layout.

A preface\index{preface} is a book's preliminary statement, usually written by the \textit{author or editor} of a work, which states its origin, scope, purpose, plan, and intended audience, and which sometimes includes afterthoughts and acknowledgments of assistance. 

When written by a person other than the author, it is called a foreword. The preface or foreword is distinct from the introduction, which deals with the subject of the work.

Customarily \textit{acknowledgments} are included as last part of the preface.

\vspace{\baselineskip}
\begin{flushright}\noindent
Place(s),\hfill {\it Firstname  Surname}\\
month year\hfill {\it Firstname  Surname}\\
\end{flushright}

%
%

\extrachap{Acknowledgements}

Use the template \emph{acknow.tex} together with the Springer document class SVMono (monograph-type books) or SVMult (edited books) if you prefer to set your acknowledgement section as a separate chapter instead of including it as last part of your preface.

\tableofcontents
%
%
%
\contributors

\begin{thecontriblist}
Firstname Surname
\at ABC Institute, 123 Prime Street, Daisy Town, NA 01234, USA, \email{smith@smith.edu}
\and
Firstname Surname
\at XYZ Institute, Technical University, Albert-Schweitzer-Str. 34, 1000 Berlin, Germany, \email{meier@tu.edu}
\end{thecontriblist}
%
%

\extrachap{Acronyms}

Use the template \emph{acronym.tex} together with the Springer document class SVMono (monograph-type books) or SVMult (edited books) to style your list(s) of abbreviations or symbols in the Springer layout.

Lists of abbreviations\index{acronyms, list of}, symbols\index{symbols, list of} and the like are easily formatted with the help of the Springer-enhanced \verb|description| environment.

\begin{description}[CABR]
\item[ABC]{Spelled-out abbreviation and definition}
\item[BABI]{Spelled-out abbreviation and definition}
\item[CABR]{Spelled-out abbreviation and definition}
\end{description}

\mainmatter
%
%
%

\begin{partbacktext}
\part{Part Title}
\noindent Use the template \emph{part.tex} together with the Springer document class SVMono (monograph-type books) or SVMult (edited books) to style your part title page and, if desired, a short introductory text (maximum one page) on its verso page in the Springer layout.

\end{partbacktext}

\title*{Modeling Speaker-Listener Interaction for Backchannel Prediction}
\author{Daniel Ortega, Sarina Meyer, Antje Schweitzer and Ngoc Thang Vu}
\institute{Institute for Natural Language Processing (IMS), University of Stuttgart, Germany \\ Email \{daniel.ortega, thang.vu\}@ims.uni-stuttgart.de}
%
%
\maketitle

\abstract*{We present our latest findings on backchannel modeling novelly motivated by the canonical use of the minimal responses \textit{Yeah} and \textit{Uh-huh} in English and their correspondent tokens in German, and the effect of encoding the speaker-listener interaction. 
Backchanneling theories emphasize the active and continuous role of the listener in the course of the conversation, their effects on the speaker's subsequent talk, and the consequent dynamic speaker-listener interaction. 
Therefore, we propose a neural-based acoustic backchannel classifier on minimal responses by processing acoustic features from the speaker speech, capturing and imitating listeners' backchanneling behavior, and encoding speaker-listener interaction. 
Our experimental results on the Switchboard and GECO datasets reveal that in almost all tested scenarios the speaker or listener behavior embeddings help the model make more accurate backchannel predictions. More importantly, a proper interaction encoding strategy, i.e., combining the speaker and listener embeddings, leads to the best performance on both datasets in terms of F1-score.}

\abstract{We present our latest findings on backchannel modeling novelly motivated by the canonical use of the minimal responses \textit{Yeah} and \textit{Uh-huh} in English and their correspondent tokens in German, and the effect of encoding the speaker-listener interaction. 
Backchanneling theories emphasize the active and continuous role of the listener in the course of the conversation, their effects on the speaker's subsequent talk, and the consequent dynamic speaker-listener interaction. 
Therefore, we propose a neural-based acoustic backchannel classifier on minimal responses by processing acoustic features from the speaker speech, capturing and imitating listeners' backchanneling behavior, and encoding speaker-listener interaction. 
Our experimental results on the Switchboard and GECO datasets reveal that in almost all tested scenarios the speaker or listener behavior embeddings help the model make more accurate backchannel predictions. More importantly, a proper interaction encoding strategy, i.e., combining the speaker and listener embeddings, leads to the best performance on both datasets in terms of F1-score. }

\section{Introduction}
\label{sec:intro}

Showing active listening plays a fundamental role in the development of any conversation and it is conveyed by signaling agreement, interest, attention, evaluation, assessment, understanding to the speaker and willingness to let them continue talking \cite{clark2004, ward2007learning, poppe2011backchannels}. These signals, referred to as \acp{bc} in the literature \cite{Yngve1970}, can appear at any time in the dialog providing feedback to the speaker and are capable to influence the course of the subsequent talk \cite{gardner1998between}. Therefore, the \ac{sli} is tied to the production and reception of \acp{bc}, which consequently lead to successful dialogs. \acp{bc} can be produced verbally, e.g. tokens like {\it Uh-huh} and {\it Yeah}, or non-verbally, e.g. nodding, gestures and smiling \cite{brunner1979, bavelas2011}. Non-verbal \acp{bc} are out of our research scope. 

\acp{bc} have been categorized under different paradigms \cite{clark2004, goodwin1986, schegloff1982discourse}. On the one hand, the lumping approach \cite{xudong2009listener, poppe2011backchannels} focuses on \ac{bc} placement, i.e. where within the speaker’s speech the \acp{bc} occur \cite{tolins2014}. On the other hand, the splitting approach aims at explaining \ac{bc} functionalities and categorizing them according to their type, function and discrete specific responses \cite{xudong2009listener}.

The minimal recipiency theory \cite{gardner1998between} addresses the canonical use of common and pervasive minimal responses in English conversation, based on the frequency with which they are found to be doing particular interactional work. Such dynamic recipiency-response interaction provides the speaker with feedback to decide how to continue with the talk. From \cite{gardner1998between}, we present a brief description of the canonical usage of \textit{Yeah} (as acknowledging response) and \textit{Uh-huh}, because our study focuses on these two specific \ac{bc} responses:

\begin{itemize}
    \item \textbf{\textit{Yeah}}  expresses that the listener aligns themself with what the speaker just uttered and claims to have understood the message. \newline
    
    \item \textbf{\textit{Uh-huh}} is the classic continuer response  \cite{schegloff1982discourse} and commonly described as example of passive recipiency \cite{jefferson1984notes} that apprises the speaker of carrying on. 
\end{itemize}

Backchannel responses are apparently present in all languages, however, their patterns differ across languages and cultures \cite{maynard1997analyzing, heinz2003}. 
Moreover, there exist other factors that can affect the \ac{bc} pattern at a personal scale, such as the personality and multilingualism of the interlocutors and the cultural and conversational and interactional context \cite{ward_2017, clancy1996, heinz2003}. These works support the importance of the interlocutors' interaction, both speakers' and listeners', and are in-line with the conversational receipt-response phenomenon presented in \cite{mcgregor2015receptionIntro}:

\begin{svgraybox}
\begin{displayquote}
\textit{The notion of recipiency is inextricably tied to the notion of response since for us \textbf{reception is response, and response is reception}. [...] simultaneous processes 
are dynamically active as a consequence of individual creativity, selectivity and/or reactivity to language use.}
\end{displayquote}
\end{svgraybox}

The \ac{bc} splitting approach, the minimal recipiency theory, and the importance of the \ac{sli} have inspired us to investigate automatic backchannel prediction from a new perspective in three different aspects that we contextualize: 

Firstly, research on computational modeling of backchanneling has mainly focused on the lumping approach,  where all \acp{bc} are considered from a single category and the task is thus seen as binary classification (absence vs presence), but differing in the technique: rule-based models \cite{ward1996using, ward2000prosodic, truong2010rule, park2017telling, al2009generating}, \ac{ml} approaches \cite{solorio2006prosodic, morency2010probabilistic} and more recently deep-learning-based models \cite{ruede2017, hara2018prediction, ishii2021multimodal, ekstedt2022voice}. 

Other works have explored the splitting approach, for instance, \cite{kawahara2016prediction} explores the automatic generation of four types of \acp{bc} in Japanese according to the dialog context, 
\cite{ortega2020oh} presents a BC predictor based on the proactive backchanneling theory (continuers vs assessments) \cite{clark2004, tolins2014}, and  \cite{blache2020integrated} introduces a backchanneler that discriminates from continuer and four specific  \ac{bc} categories. We align with the latter works by following the splitting approach and exploring \textit{Yeah} and \textit{Uh-huh} in their canonical usage as \acp{bc}. 

Secondly, because of the importance of the interlocutors' interaction, the cultural context and the individual facts that shape the conversation course, we hypothesize that encoding the speaker and listener behaviors, to be thereafter combined to model the \ac{sli}, can also have a positive impact on the overall model's performance. \cite{ortega2020oh} reported that encoding the listener behavior helps the automatic \ac{bc} prediction. This supports our hypothesis that interlocutors' and interactional cues can contribute to make better predictions, if they are properly encoded. 

Thirdly, we believe that modeling the interlocutors' interaction in the context of minimal responses is a form of dealing with the conversational receipt-response phenomenon \cite{mcgregor2015receptionIntro} previously quoted. 

Motivated by the splitting approach, the minimal recipiency theory and previous works, we propose a novel neural backchannel predictor. Based on acoustic features from the speaker speech, our model detects BC opportunities for \textit{Yeah} and \textit{Uh-huh} in their canonical usage as BC responses. Additionally, our model predicts \textit{No-BC} (no backchannel) for speaker speech regions that do not elicit any BC response, in other words, \textit{No-BC} represents BC absence.

To the best of our knowledge, \ac{bc} modeling has never been approached from this angle.  
Furthermore, no corpus exists that is manually annotated at such fine-grained categories. Therefore, a semi-automatic and heuristic annotation was performed on two corpora of dialogs with manual transcriptions: \ac{swda} \cite{jurafsky1997switchboard} in English and the \ac{geco} \cite{schweitzer2013conv}. 
Moreover, we present three different methods to encode the speaker-listener interaction: sum, bilinear transformation and \ac{ntn}. All of them aim to encode the relationship between the speaker and listener embeddings. 

Our experimental results on our corpora confirm that by encoding either the listener or the speaker behavior the model performance improves consistently in terms of F1-score and more importantly encoding the speaker-listener interaction leads to the best performance on both datasets.  In order to ensure reproducibility, our code, annotation and data will be publicly available in our repository\footnote{\url{https://github.com/DigitalPhonetics}}.



\section{Related work}
\label{sec:related_work}

Automatic \ac{bc} prediction has been approached in the past few years as a classification task. In this section, we enumerate a wide variety of works, that cover different theories, methods, types of BCs, data, and model input features, e.g. acoustic, lexical and visual from the speaker. Therefore, it is difficult to directly compare the models and their results. This literature review serves to situate the state of the art and place our work and its novelties in the field.

The literature shows a large number of \ac{bc} predictors based on rules. For instance, \cite{ward1996using, ward2000prosodic, gravano09_interspeech, truong2010rule, park2017telling} and \cite{truong2010rule} consider a subset of prosodic features such as speech regions with low or rising pitch, high or decreasing energy patterns, pausal information.
\cite{al2009generating} expands the scope by presenting a multimodal rule-based model using prosodic and facial features. 


\ac{ml} models for automatic prediction have more recently been built, ranging from classical approaches to neural-based ones. \cite{solorio2006prosodic} proposed an algorithm based on logically weighted linear regression showing similar results at the time compared to rule-based models. \cite{morency2010probabilistic} implemented a multi-modal approaches, where lexical, prosodic and visual (eye gaze) cues were employed as features for sequential probabilistic models, i.e. conditional random fields and hidden Markov models. 

With respect to \ac{dl} models, \cite{ruede2017} and \cite{hara2018prediction} explored the \ac{bc} prediction using \acp{lstm} and \acp{ffn}. The former employed prosodic and \acp{mfcc} and lexical features, while the latter only prosodic. \cite{ekstedt2022voice}
explored turn-taking events, including \ac{bc} prediction using transformer-based model over raw speech signals. \cite{hara2018prediction} proposed a multitask learning approach by jointly learning and predicting \acp{bc}, turn-taking and fillers.  Another multi-modal multitask work is in \cite{ishii2021multimodal}, lexical features are processed using BERT-based transformers  \cite{devlin2018bert}. \cite{jain2021exploring} proposed a semi-supervised model to predict backchannel opportunities for verbal, visual and combined \acp{bc} using a multimodal (audio and video) \ac{lstm} fusion based architecture.

The aforementioned predictors have only considered backchanneling as a binary decision (absence vs presence), a lumping approach \cite{xudong2009listener, poppe2011backchannels} where all types of \acp{bc} are considered as a single one and the focus is on finding right time opportunities for a \acp{bc} in the speaker's talk.

On the other hand, the splitting approach \cite{xudong2009listener} examines discrete and specific \ac{bc} responses. In alignment to the splitting approach, some predictors have taken the task to a more fined \ac{bc} granularity. For instance, 
\cite{kawahara2016prediction} investigated the automatic generation of four types of Japanese \acp{bc} according to the dialog context, using linguistic features preceding \acp{bc}, part of speech tags, and prosodic features as input for a logistic regression model.
\cite{ortega2020oh} built a \ac{cnn}-based model following the proactive backchanneling theory \cite{clark2004, tolins2014} differentiating \acp{bc} between generic/continuers and specific/assessments.
\cite{blache2020integrated} introduced a bimodal (verbal and visual) backchanneler that discriminates from generic and four specific \ac{bc} categories.

Our model also falls in the splitting approach. However, its novelty resides in its ability to predict \textit{Yeah} and \textit{Uh-huh} in their canonical use as BCs, following the minimal recipiency theory. Moreover, based on the models, experiments and positives results from \cite{ruede2017} and \cite{ortega2020oh}, our model takes \acp{mfcc} features as input. 

\begin{figure}
    \caption{Backchannel prediction's model with speaker-listener interaction. Our model is depicted with four constituents: acoustic component, speaker and listener embedding modules and the speaker-listener encoding mechanism. Not all constituents interact simultaneously, but different variants are introduced and described in \autoref{sec:method} along with their outcome in \autoref{ssec:exp_results}. $\oplus$ represents a concatenation.}
    \centering
    \includegraphics[width=\textwidth,trim={0 0 0 -50},clip]{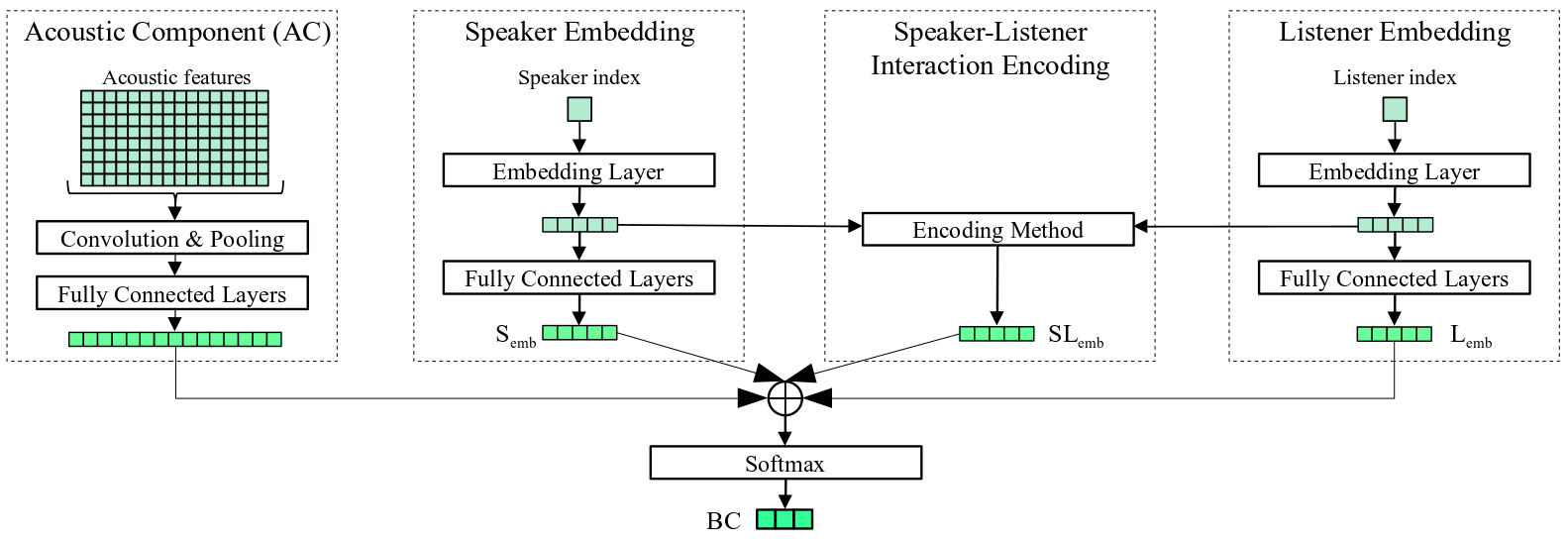}
    \label{fig:model} 
\end{figure}

\section{Proposed model}
\label{sec:method}

We introduce a \ac{bc} predictor, depicted in \autoref{fig:model}, that consists of an \acf{ac} and three behavior encoders, whose outputs are concatenated for posterior classification into \textit{No-BC}, \textit{Yeah}, or \textit{Uh-huh}: \newline

\noindent\textbf{\acs{ac}:}  An acoustic \ac{cnn} that generates a vector representation from the frame-level \acp{mfcc} features of the last 2000 ms given a time point that are extracted from the speaker channel. This approach is based on \cite{ortega2019}. \acp{cnn} perform a discrete convolution using a set of different filters on an input matrix $w$, where each column corresponds to the frame-level \acp{mfcc}.  We use 2D filters $f$ (with width $|f|$, i.e. number of consecutive frames) spanning over all dimensions $d$, i.e. number of \ac{mfcc} features, as described by \autoref{eq:convolution}.
    \begin{equation}\label{eq:convolution}
	    \small (w \ast f)(x,y) = \sum_{i=1}^{d}\sum_{j = -|f|/2}^{|f|/2}w(i,j) \cdot f(x-i,y-j)
    \end{equation}   
\noindent After convolution, the output is passed though a non-linear function, ReLU in this case, followed by a max-pooling operation in order to extract and concatenate the highest activation per filter, also known as feature map. \newline

\noindent\textbf{Listener embedding:} It is an embedding layer, based on \cite{ortega2019}, followed by two \acp{ffn}. The embedding layer is a type of neural hidden layer that learns, encodes and stores dense vector representations of the listeners' \ac{bc} patterns in relation with the model input. \newline

\noindent\textbf{Speaker embedding:} An embedding layer that encodes the speaker's pattern. The mechanism is identical to the listener embedding. \newline
    
\noindent\textbf{\ac{sli}:} encodes the speaker-listener interaction. We propose in \autoref{ssec:emb_encoding} three speaker-listener interaction encoding mechanisms. \newline

The \ac{ac} is always present in the model, but not all behavior encoders interact simultaneously. Different variants are experimented and reported in \autoref{ssec:exp_results}. 

Finally, the feature map and three resulting vectors from the behavior encoders, depending on the variant, are concatenated and passed to a softmax layer, that outputs a probability distribution over the three classes: \textit{No-BC}, \textit{Yeah} or \textit{Uh-huh}.

\subsection{Encoding speaker-listener interaction}
\label{ssec:emb_encoding}

Our hypothesis is that the interlocutors’ interaction can be modeled, i.e. encoding jointly the speaker and listener behaviors, and yield a positive impact on the overall model’s performance.  Therefore, the three methods for encoding \ac{sli} presented below were implemented and tested during the experimentation phase. We follow this notation: $S_{emb}$ stands for speaker embedding, $L_{emb}$ for listener embedding and $SL_{emb}$ for speaker-listener interaction embedding. $S_{emb}$ and $L_{emb}$ are coming from two independent embedding layers.
\newline

\noindent\textbf{Sum:} This is the most simple approach and consists of summing up the interlocutor's embeddings as in \autoref{eq:sum}.
    \begin{equation}\label{eq:sum}
    \small SL_{emb} = S_{emb} + L_{emb}
    \end{equation}

\noindent\textbf{Bilinear:} A bilinear tranformation is applied to $S_{emb}$ and $L_{emb}$ and defined in \autoref{eq:bilinear}:    
    \begin{equation}\label{eq:bilinear}
    \small SL_{emb} = S_{emb}^T \; W \; L_{emb} + b
    \end{equation}
    
 \noindent where $W \in \mathcal{R}^{d \times d \times k}$ stands for the learnable weight matrix, $n$ is the dimension of $S_{emb}$ and $L_{emb}$ and $k$ the dimension of  $SL_{emb}$, and $b$ the bias term.
\newline

\noindent    
\textbf{\ac{ntn}:} The goal of the \ac{ntn} is to find and encode the relationship between two entity vectors across multiple dimensions \cite{socher2013reasoning, ding2015deep}, by computing a score ($SL_{emb}$) that defines how likely it is that the two entities are in a particular relationship. It is defined in \autoref{eq:ntn}:
    \begin{equation}\label{eq:ntn}
        \small SL_{emb} = U^T  \tanh\Bigl(S^T_{emb} \: W^{[1:k]} \: L_{emb} + V \begin{bmatrix}
            S_{emb}\\
            L_{emb} 
            \end{bmatrix} \Bigr) + b
    \end{equation}

 \noindent where $tanh$ is applied element-wise, $W^{[1:k]} \in \mathcal{R}^{n \times n \times k}$ is a learnable weight matrix and $S^T_{emb} \: W^{[1:k]} \: L_{emb}$ stands for a bilinear transformation, already introduced in \autoref{eq:bilinear}, but applied $k$ times, each entry takes a slice of $W$. $V\in \mathcal{R}^{k \times 2d}$, $U \in \mathcal{R}^{k}$ and $b$ (bias) are learnable parameters \cite{socher2013reasoning}.


\section{Experimental setup}
\label{sec:experimental_setup}

We experimented on two corpora: \ac{swda} for English and \ac{geco} for German. We present their specs and statistics on \autoref{tab:data_stats}. Our data annotations, described below, are available in the project repository.

\begin{SCtable}[]
\footnotesize
\caption{Statistics for SwDA and GECO datasets and their class distribution.}
\label{tab:data_stats}
	\begin{tabular}{lcccc}
		\hline\noalign{\smallskip}
		\multicolumn{1}{c}{\multirow{2}{*}{\diagbox{Spec}{Dataset}}}& \multicolumn{2}{c}{ SwDA}& \multicolumn{2}{c}{ GECO}\\
		\multicolumn{1}{c}{}	& Counts \thickspace & \% \thickspace\thickspace & Counts  \thickspace	& \% \\
        \noalign{\smallskip}\svhline\noalign{\smallskip}
        \textit{Yeah }	    &15,380	&19.3 &2,026	&24.3 \\
		\textit{Uh-huh} 	    &24,436	&30.7  &2,149	&25.7 \\
		\textit{No-\ac{bc}} &39,816	&50.0  &4,175	& 50.0 \\
		Conversations   &2438 &--  &46	& -- \\
		Interlocutors	        &520 &	--   &13  &-- \\
		\noalign{\smallskip}\hline\noalign{\smallskip}
	\end{tabular}
\end{SCtable}

\subsection{The Switchboard Dialog Act Corpus}
\label{sssec:swbd}

\ac{swda} \cite{jurafsky1997switchboard} is a dialog corpus of telephone English conversations annotated at dialog-act level, including \acp{bc} . 
Annotations and time stamps for general BCs are from  \cite{jurafsky1997switchboard}, while for \textit{No-BC} instances, splits and time stamps are taken from \cite{ruede2017}. According to the authors, the latter time stamps were chosen from a few seconds of speech before each BC from the speaker speech to constitute negative samples (BC absence) and they motivate it by assuming that during those periods the listener decided not to backchannel regardless of what the speaker was saying. With respect to the splits, from a total of 2,438 conversations, 2,000 are for training, 200 for validation and 238 for testing.
We used the categorization proposed and mentioned in \autoref{sec:intro}, i.e recipiency tokens used for backchanneling \textit{Yeah} vs \textit{Uh-huh}. For that purpose, we manually annotated the \ac{bc} tokens and kept those that fitted into both categories. 

Our annotation process for \ac{bc} was as follows, we extracted a list of 670 unique utterances used within the corpus to backchannel. In the category \textit{Uh-huh}, we included all \ac{bc} realizations like {\it uh-huh}, {\it um-hum} and other variants, that describe passive recipiency. 
For the second category \textit{Yeah}, we included all \ac{bc} realizations like {\it yeah}, {\it yes}, \textit{yep} and other variants, that express alignment to the speaker. We excluded variants with the marker \textit{[laughter]}, because the laughter assesses the speaker's talk and does not meet the canonical use of these minimal responses that we investigate. Any other \ac{bc} realization, marker or combination was discarded. 

During the annotation process, 84 unique \ac{bc} realizations out of 670 were finally included, 57 for the category \textit{Yeah} and 27 for \textit{Uh-huh}. As result, our data consists of 15.4k \textit{Yeah} instances and 22.4k of \textit{Uh-huh}, summing up to 39.8k \ac{bc} instances and the same amount for \textit{No-\ac{bc}}.

For speaker/listener annotation, \ac{swda} provides the mapping between dialog channels and unique interlocutors. On average, each speaker takes part in 10 conversations. Official documentation reports 543 unique participants, but the annotation does not include the speaker ID in 11 channels. Therefore, we followed \cite{ortega2020oh}'s suggestion and a random speaker was assigned to those channels, ending up with 520 unique interlocutor IDs.

\subsection{GErman COrpus database}
\label{sssec:geco}

\ac{geco}  \cite{schweitzer2013conv} consists of 46 two-party dialogs in German between unacquainted female subjects. 22 conversations took place in a unimodal setting, where participants could not see each other, while for the remaining 24 dialogs subjects were facing each other. 

\ac{bc} annotation was done heuristically based on acoustic and lexical information, because no manual annotation was available. An utterance expressed during another person's turn is regarded as \ac{bc} if its duration is less than one second, or if all its words are covered in a manually generated list of German \ac{bc} expressions. The list contains common single-word \ac{bc} like \textit{ja} (\textit{yes}) and \textit{hm}, multi-word expressions like \textit{oh mein Gott} (\textit{oh my god}), and intensifiers for \ac{bc} such as \textit{voll} (\textit{very}) in \textit{voll cool} (\textit{really cool}).

During the initial annotation process, 1,657 utterances were marked as unique \ac{bc} realizations. However, after manual checking, 1,005 were considered \acp{bc} and the rest false positives. For \textit{Yeah} vs \textit{Uh-huh}, we followed the same annotation procedure explained in \autoref{sssec:swbd} . For the former, we included the terms \textit{ja} and other variants, while for the latter  \textit{hm hm}, \textit{mh mh} and \textit{mhm mhm}. Finally, 91 unique \ac{bc} realizations were included in the dataset, 37 for the category \textit{Yeah} and 54 for \textit{Uh-huh}. As result, our data consists of 2k \textit{Yeah} instances and 2.1k of \textit{Uh-huh}, summing up to 4.1k \ac{bc} instances and the same amount for \textit{No-\ac{bc}}. We followed the method from \cite{ruede2017}, described in \autoref{sssec:swbd}, to select \textit{No-BC} instances. 


\subsection{Acoustic features extraction}
\label{sssec:ac_data}

As mentioned in \autoref{sssec:swbd} and \autoref{sssec:geco}, we obtained the time stamps for all instances in the dataset that were later used to extract acoustic features of the 2000 ms from the speaker speech signal before the \ac{bc} happens. 

\cite{ortega2020oh} experimented with \acp{mfcc} and prosodic features on \ac{bc} modeling and found that \acp{mfcc} features yield better results. Therefore, we also consider 13 \acp{mfcc} features that were extracted using the openSMILE toolkit \cite{tools:openSMILE} at frame level, i.e. the speech signal is divided into frames of 25 ms with a shift of 10 ms.

\section{Experimental results}
\label{ssec:exp_results}

We present the results of model variants trained on both datasets. All variants include the \ac{ac} depicted in  \autoref{fig:model}, and they differ by the components that are concatenated, i.e. speaker embedding (S), listening embedding (L) and \ac{sli} embedding. The hyperparameter ranges used for experimentation are presented in \autoref{tab:hyperparams}. The listener and speaker embeddings are 5-dim vectors.

\begin{SCtable}[][ht]
\footnotesize
\caption{ Range values used for the hyperparameter search during the model training and fixed values for the most relevant hyperparameters of the model architecture. }
\label{tab:hyperparams}
\begin{tabular}{ll}
\hline\noalign{\smallskip}
Hyperparameter 	 &Value\\ 
\noalign{\smallskip}\svhline\noalign{\smallskip}
    Filter widths 				& [10, 11, 12]\\
    Number of filters           &[16, 32, 64, 128] \\
    Dropout rate			    &[0.1, 0.3, 0.5] \\
    Pooling size		 	    &(10, 1)\\
    Acoustic features		    & MFCC \\
    Number of speech frames\thickspace\thickspace			&[48, 98, 148, 198]\\
    Mini-batch size				&[16, 32, 64, 128]  \\
    Embedding length            &5  \\
\noalign{\smallskip}\hline\noalign{\smallskip}
\end{tabular}
\end{SCtable}

Our first experiments aimed to explore the performance of the \ac{ac}, our baseline, and the effect of concatenating the embeddings.  Results are shown in the upper part of \autoref{tab:experimental_results}. They were obtained on the test set when the best performance was found on the validation set. On \ac{swda}, the \ac{ac} reached an 56.9\% in accuracy and 0.42 in F1-score. By concatenating the speaker embedding (AC $\oplus$ S), both the accuracy and F1-score show a slight improvement. Whereas by concatenating the listener embedding (AC $\oplus$ L), the metrics improve even further, reaching 60.3\% and 0.52 in accuracy and F1-score, accordingly. Finally, we concatenated both embeddings (AC $\oplus$ S $\oplus$ L) and a slight extra boost was reached in both metrics, bringing an improvement of 3.8\% in accuracy and 0.11 in F1-score in comparison with the baseline.

On \ac{geco}, the trend of the results is slightly different. The \ac{ac} reached an 49.3\% in accuracy terms and 0.35 in F1-score. The accuracy is not even reaching 50\%, i.e. the majority class \textit{No-BC} in our setup. Nonetheless, by concatenating the speaker embedding, both metrics improved, the accuracy reaching 55.1\% and the F1-score 0.42. We take this scenario as our baseline on this dataset. Unlike the experiments on \ac{swda}, when \ac{ac} and the listener embeddings interact, the performance drops drastically. Finally, we concatenated both embeddings and as outcome both metrics improved in comparison with the baseline, reaching  60.0\% in accuracy and 0.44 in F1-score. 

\begin{SCtable}[][ht]
\footnotesize
\caption{Accuracy (\%) and F1-score for \ac{ac} (row 1), its variants with concatenated embeddings (rows 2-4) and with methods encoding the speaker-listener interaction  (lower part, rows 5-7). $\oplus$ stands for concatenation.}	
\label{tab:experimental_results} 
		\begin{tabular}{lcccc}
			\hline\noalign{\smallskip}
			\multicolumn{1}{c}{\multirow{2}{*}{Model}}& \multicolumn{2}{c}{ \ac{swda}}& \multicolumn{2}{c}{GECO}\\
			\multicolumn{1}{c}{}	&Accuracy \thickspace	 & F1-score \thickspace & Accuracy \thickspace 	& F1-score\\
            \noalign{\smallskip}\svhline\noalign{\smallskip}
            Acoustic (AC)	    &56.9	&0.42   &49.3	&0.35 \\
			AC $\oplus$ Speaker (S)	&59.7	&0.46   &55.1	&0.42 \\
			AC $\oplus$ Listener (L)   &60.3   &0.52   &44.9	&0.38 \\
			AC $\oplus$ S $\oplus$ L	        &\textbf{60.7}	&\textbf{0.53}   &\textbf{60.0}   &\textbf{0.44} \\
			\noalign{\smallskip}\hline\noalign{\smallskip}
			AC $\oplus$ \ac{sli}-Sum        &60.5   &0.54   &62.7	&0.49  \\
			AC $\oplus$ \ac{sli}-Bilinear	&58.8	&0.50   &58.2	&0.37 \\
            AC $\oplus$ \ac{sli}-NTN	    &\textbf{61.0}	&\textbf{0.55}   & \textbf{64.7}	& \textbf{0.51} \\
            \noalign{\smallskip}\hline\noalign{\smallskip}
		\end{tabular}

\end{SCtable}

\subsection{Effect of encoding mechanisms}
\label{ssec:enc_mecs}
We found that the embeddings let the model make more accurate predictions. Therefore, we experimented with the three mechanisms to encode the \ac{sli} from \autoref{ssec:emb_encoding}: Sum, Bilinear and \ac{ntn}. Our intuition behind this was that a more sophisticated mechanism could capture better the interaction between the interlocutors, and ultimately it would improve the \ac{bc} prediction.  The lower part of \autoref{tab:experimental_results} contains these results. 

We took the model variant AC$\oplus$S$\oplus$L as baseline to be compared with the \ac{sli} encoding mechanisms. On \ac{swda}, the embedding Sum does not provide any improvement. Moreover, the Bilinear approach even affects the performance. Finally, the \ac{ntn} showed a slight boost on both metrics, 61.0\% for accuracy and 0.55 for F1-score. 
On \ac{geco}, the embedding Sum enhances the model performance to 62.7\% on accuracy and 0.49 on F1-score, whereas the Bilinear mechanism exerts negative influence on both metrics with respect to the current baseline. At last, the \ac{ntn} surpassed the rest of the previous mechanisms ending up at 64.7\% on accuracy and 0.51 in F1-score.

On both datasets, the Bilinear approach affected negatively the performance, showing that it is not suitable for encoding the \ac{sli}. The Sum mechanism helped on \ac{geco}, but not on \ac{swda}, leading to no conclusive interpretation. Conversely, the NTN mechanism substantially enhanced the model performance.

Although considerable improvements on accuracy can be consistently seen during the experiments, the best performing model on both datasets is still far from solving the task. We thus inquire deeper into the model results in \autoref{ssec:class_distrbution}.

\subsection{Effect of class distribution}
\label{ssec:class_distrbution}

The class distribution is skewed and the accuracy analysis is not enough. Therefore, we look closer at macro F1-scores and per class.  On the one hand, the improvement is notable when we compare macro F1-scores from \ac{ac} (baseline) vs AC$\oplus$\ac{sli}-NTN (best performing model) on both datasets, see \autoref{tab:experimental_results}. On the other hand, when we look at the F1-score per class from the model AC$\oplus$\ac{sli}-NTN on both datasets, the dominance of the \textit{No-BC} class becomes evident, see \autoref{tab:f1_score_per_class}. 

\begin{SCtable}[][ht]
\footnotesize
\caption{F1-scores per class on both datasets from the model variant: AC$\oplus$\ac{sli}-NTN.}
\label{tab:f1_score_per_class} 
	\begin{tabular}{llcccc}
		\hline\noalign{\smallskip}
		& \multicolumn{1}{c}{\multirow{2}{*}{Dataset}}& \multicolumn{3}{c}{ F1-score} & \\
		&	& \thickspace\textit{No-BC}	& \thickspace\textit{Yeah} & \thickspace\textit{Uh-huh} & \\
        \noalign{\smallskip}\svhline\noalign{\smallskip}
        & \ac{swda}	&0.70	&0.35  &0.58 &  \\
		& \ac{geco}	&0.78	&0.18  &0.56 &  \\
		 \noalign{\smallskip}\hline\noalign{\smallskip}
	\end{tabular}
\end{SCtable}

The model performs fairly well on classes \textit{No-BC} and \textit{Uh-huh}, but struggles with the category \textit{Yeah}. We hypothesize two possible reasons: (1) the class \textit{Yeah} is the smallest in the dataset, and (2) the elicitation of acknowledging responses, like \textit{Yeah}, depends more on lexical and semantic information \cite{gardner1998between} that are not currently considered in this investigation scope. The integration of such information and the mechanism to process it are left as future work.

\section{Exploring the listeners embeddings}
\label{sec:expl_lis_emb}

\label{ssec:Listeners_embeddings}

We observed the contributions of encoding the \ac{sli}. Therefore, we decided to experiment and analyze even further the trained interlocutors' embeddings, especially the listeners embeddings from two model variants: AC$\oplus$L (best performing model without interaction encoding) and AC$\oplus$\ac{sli}-NTN (best performing model with interaction encoding). The former model only learns listener embeddings, while the latter model learns jointly listener and speaker embeddings.


\subsection{Impact of listener embeddings}
\label{ssec:impact_list_embs}
\begin{figure}[!htb] 
\sidecaption
    \centering
    \includegraphics[width=0.60\textwidth,trim={0 10 0 25},clip]{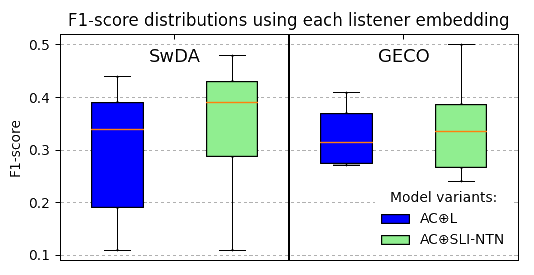}
    \caption{F1-score distributions for predicting BC using each listener embedding separately on \ac{swda} (left side) and \ac{geco} (right side). Distributions in blue (darker) correspond to AC$\oplus$L, whereas distribution in green (lighter) to AC$\oplus$\ac{sli}-NTN.}
    \label{fig:boxplots}
\end{figure}

Given that each listener embedding encodes the \ac{bc} behavior, we wanted to know at which extent a particular embedding can help the model to accurately predict \acp{bc}, even in conversations where this embedding did not originally take part. To accomplish this goal, the model made predictions for the whole test set using only one of listener embeddings at a time and we calculated the corresponding F1-score. This series of predictions results in four F1-score distributions, one per model variant vs. dataset, depicted in  \autoref{fig:boxplots}.

From the F1-score distributions on \ac{swda}, we found that the model AC$\oplus$L achieved  the highest F1-score of 0.44 (mean=0.29, median=0.34), whereas the model AC$\oplus$\ac{sli}-NTN  achieved the highest F1-score of 0.48 (mean=0.35, median=0.39). Furthermore, by directly comparing listeners on both scenarios, 437 listeners out of 520 (84.0\%) led to an improvement in terms of F1-score, 0.06 on average, but within the interval [-0.14, 0.28].



Similar findings come from the F1-score distribution on \ac{geco}, although the number of unique interlocutors is comparably smaller. While the highest F1-score from the model AC$\oplus$L was 0.40 (mean=0.33, median=0.32), the model AC$\oplus$\ac{sli}-NTN scored 0.50 (mean=0.34, median=0.32) as maximum.  
Finally, seven listeners out of 13 showed an improvement of 0.02 in F1-score on average, within the interval [-0.14, 0.16]. 

\begin{figure*}[!htb] 
  \begin{minipage}[b]{0.48\linewidth}
    \centering
    \caption{PCA-reduced listeners embeddings distribution from the model variant AC$\oplus$L on  \ac{swda}. Axes correspond to the the principal components. Centroid appears as a green cross. The color of each dot represents the F1-score according to the bar on the right.\\}
    \includegraphics[width=\textwidth,trim={27 30 35 33},clip]{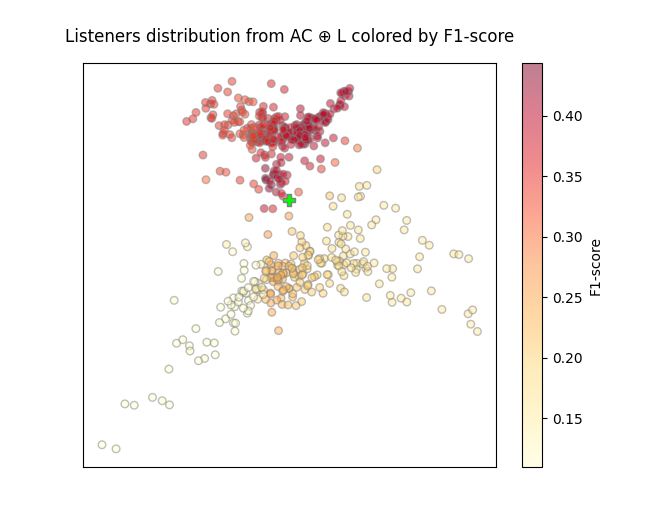}
    
    \label{fig:SWDB_L_Emb_BI}
    \vspace{4ex}
  \end{minipage}
  \hspace{3ex}
  \begin{minipage}[b]{0.48\linewidth}
    \centering
    \caption{PCA-reduced  listeners embeddings distribution from the model variant AC$\oplus$\ac{sli}-NTN on \ac{swda}. Axes correspond to the the principal components. Centroid appears as a green cross. The color of each dot represents the F1-score according to the bar on the right. \\}
    \includegraphics[width=\textwidth,trim={27 30 35 33},clip]{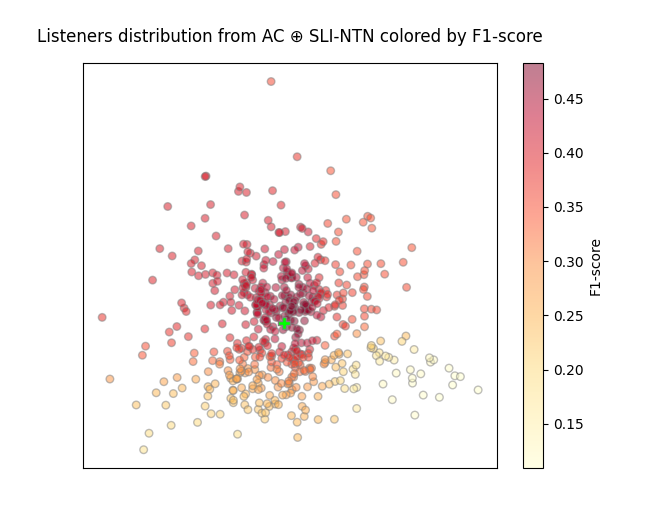}
    
    \label{fig:SWDB_L_Emb_AI}
    \vspace{4ex}
  \end{minipage} 
  
  \begin{minipage}[b]{0.48\linewidth}
    \centering
    \caption{Listeners frequency distribution obtained from the model variant AC$\oplus$L on \ac{swda}. Mean and median are depicted numerically and with vertical lines, blue and green, respectively.\\}
    \includegraphics[width=\textwidth,trim={0 10 0 30},clip]{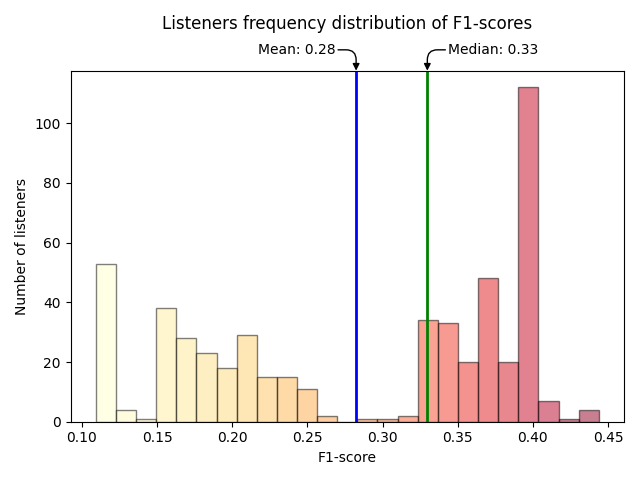}
    \label{fig:SWDB_L_Emb_BI_hist} 
  \end{minipage}
  \hspace{3ex}
  \begin{minipage}[b]{0.48\linewidth}
    \centering
    \caption{Listeners frequency distribution obtained from the model variant AC$\oplus$\ac{sli}-NTN on \ac{swda}. Mean and median are depicted numerically and with vertical lines, blue and green, respectively.\\}
    \includegraphics[width=\textwidth,trim={0 10 0 30},clip]{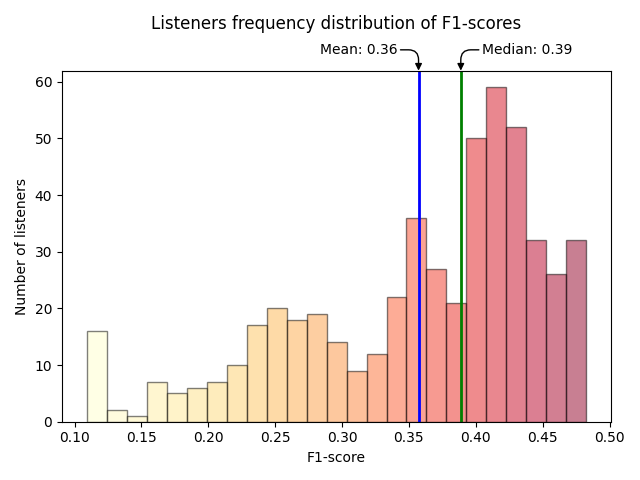}
    \label{fig:SWDB_L_Emb_AI_hist}
  \end{minipage} 
\end{figure*}

\subsection{A closer look at the embedding space}

We inspected the listeners embeddings and the F1-score distributions from both models variants analyzed in previous \autoref{ssec:impact_list_embs}. 
Nevertheless, we restricted  our investigation to \ac{swda}, because of the vast number of unique listeners. 

Initially, we extracted the listeners embeddings from each model variant and applied \ac{pca} to reduce the embedding dimensionality from five to two dimensions, in order to be 2D-plotted as listener distributions. They can be seen on \autoref{fig:SWDB_L_Emb_BI} and \autoref{fig:SWDB_L_Emb_AI}. The former corresponds to the model AC$\oplus$L, while the latter to AC$\oplus$\ac{sli}-NTN. Each dot represents one speaker embedding and its color intensity the F1-score, according to the color scale on the right. The green cross represents the centroid of the distribution.

\autoref{fig:SWDB_L_Emb_BI_hist} and \autoref{fig:SWDB_L_Emb_AI_hist} depict the frequency distribution of F1-score in 25 intervals
, corresponding to the model variant AC$\oplus$L and AC$\oplus$\ac{sli}-NTN, respectively, including the mean (blue line) and the median (green line).



With respect to the listeners' embedding distribution from AC$\oplus$L, \autoref{fig:SWDB_L_Emb_BI}, we can observe two clear clusters, one above (redder/darker) and one below (more yellow/lighter) the centroid. The upper cluster concentrates the majority of listeners embeddings that help the model variant AC$\oplus$L achieve the highest F1-scores. It is important to remark that embeddings do not tend to concentrate around the centroid. The frequency distribution on \autoref{fig:SWDB_L_Emb_BI_hist} also shows the two aforementioned clusters, with few embeddings in the middle. 


Regarding with the listeners' embedding distribution from AC$\oplus$\ac{sli}-NTN, \autoref{fig:SWDB_L_Emb_AI}, it behaves differently. The embeddings cluster more uniformly and the reddest/darkest areas concentrate around the centroid. Nonetheless, the upper area groups the the majority of listeners embeddings that help the model variant AC$\oplus$\ac{sli}-NTN achieve the highest F1-scores. The frequency distribution from \autoref{fig:SWDB_L_Emb_AI_hist} shows that most of the listener embeddings accumulate on the right side, the area containing the higher F1-scores.



As future work, we consider a more extensive analysis that includes the relation between characteristics linked to a specific listener, e.g. the F1-score vs the number of \acp{bc} in the training set and the number of conversations with unique speakers.

\section{Conclusions}


In this paper, we presented a novel approach for backchannel prediction characterized by three aspects: 1) it is motivated by minimal recipiency theory, i.e. the canonical use of minimal responses \textit{Yeah} and \textit{Uh-huh} in both English and German, 2) it encodes the speaker and listener behavior, and 3) it models the speaker-listener interaction performing three different mechanisms.  
Additionally, we implemented a semi-automatic and heuristic annotation on the corpora \ac{swda} and \ac{geco} that is publicly available for further research. 

Our experimental results on both datasets showed that by combining the acoustic component with either the listener or the speaker embeddings, the model performance steadily improves. More importantly, the speaker-listener interaction helped the model reach its best overall performance. Finally, by exploring and experimenting with the listeners embeddings from two model variants on \ac{swda}, we found that 84.0\% of the listener embeddings helped the model to make more asserted predictions, when they were learnt jointly with the speaker embeddings. 




\bibliography{custom}
\bibliographystyle{spmpsci}

\begin{acronym}[Bash]
    \acro{adagrad}[AdaGrad]{adaptive gradient algorithm}
    \acro{am}[AM]{attention mechanism}
    \acro{crf}[CRF]{conditional random field}
    \acro{cnn}[CNN]{convolutional neural network}
    \acro{lstm}[LSTM]{long short-term network}
    \acro{ffn}[FFN]{feed-forward network}
    \acro{da}[DA]{dialog act}
    \acro{dl}[DL]{deep learning}
    \acro{e2e}[E2E]{End-to-End}
    \acro{gd}[SGD]{stochastic gradient descent}
    \acro{hmm}[HMM]{hidden Markov model}
    \acro{nlp}[NLP]{natural language processing}
    \acro{rnn}[RNN]{recurrent neural network}
    \acro{svm}[SVM]{support vector machine}
    \acro{swbd}[SWBD]{Switchboard Corpus}
    \acro{swda}[SwDA]{The Switchboard Dialog Act Corpus}
    \acro{asr}[ASR]{automatic speech recognition}
    \acro{nn}[NN]{neural network}
    \acro{mt}[MTs]{manual transcriptions}
    \acro{at}[ATs]{automatic transcriptions}
    \acro{e2e}[E2E]{End-to-End}
    \acro{cd}[CD]{context-dependent}
    \acro{tdnn}[TDNN]{time-delay neural network} 
    \acro{ctc}[CTC]{connectionist temporal classification}
    \acro{wer}[WER]{word error rate}
    \acro{mfcc}[MFCC]{Mel-frequency cepstral coefficient}
    \acro{gmm}[GMM]{Gaussian Mixture Model}
    \acro{espnet}[ESPnet]{End-to-End Speech Processing Toolkit}
    \acro{bc}[BC]{backchannel}
    \acro{geco}[GECO]{GErman COrpus}
    \acro{ntn}[NTN]{Neural Tensor Network}
    \acro{ac}[AC]{Acoustic Component}
    \acro{pca}[PCA]{principal component analysis}
    \acro{sli}[SLI]{speaker-listener interaction}
    \acro{ml}[ML]{machine learning}
\end{acronym}

\backmatter
\appendix
%
%
%

\chapter{Chapter Heading}
\label{introA} 

Use the template \emph{appendix.tex} together with the Springer document class SVMono (monograph-type books) or SVMult (edited books) to style appendix of your book in the Springer layout.

\section{Section Heading}
\label{sec:A1}
Instead of simply listing headings of different levels we recommend to let every heading be followed by at least a short passage of text. Further on please use the \LaTeX\ automatism for all your cross-references and citations.

\subsection{Subsection Heading}
\label{sec:A2}
Instead of simply listing headings of different levels we recommend to let every heading be followed by at least a short passage of text. Further on please use the \LaTeX\ automatism for all your cross-references and citations as has already been described in Sect.~\ref{sec:A1}.

For multiline equations we recommend to use the \verb|eqnarray| environment.
\begin{eqnarray}
\vec{a}\times\vec{b}=\vec{c} \nonumber\\
\vec{a}\times\vec{b}=\vec{c}
\label{eq:A01}
\end{eqnarray}

\subsubsection{Subsubsection Heading}
Instead of simply listing headings of different levels we recommend to let every heading be followed by at least a short passage of text. Further on please use the \LaTeX\ automatism for all your cross-references and citations as has already been described in Sect.~\ref{sec:A2}.

Please note that the first line of text that follows a heading is not indented, whereas the first lines of all subsequent paragraphs are.

%
\begin{figure}[t]
\sidecaption[t]
\includegraphics[scale=.65]{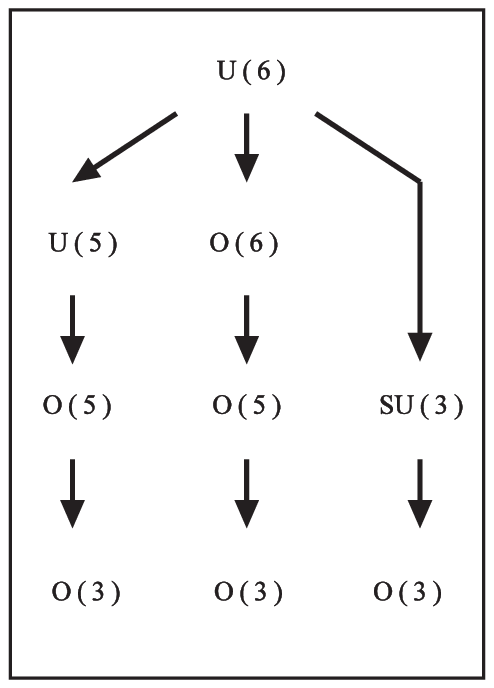}
%
%
\caption{Please write your figure caption here}
\label{fig:A1}       
\end{figure}

%
\begin{table}
\caption{Please write your table caption here}
\label{tab:A1}       
%
%
\begin{tabular}{p{2cm}p{2.4cm}p{2cm}p{4.9cm}}
\hline\noalign{\smallskip}
Classes & Subclass & Length & Action Mechanism  \\
\noalign{\smallskip}\hline\noalign{\smallskip}
Translation & mRNA$^a$  & 22 (19--25) & Translation repression, mRNA cleavage\\
Translation & mRNA cleavage & 21 & mRNA cleavage\\
Translation & mRNA  & 21--22 & mRNA cleavage\\
Translation & mRNA  & 24--26 & Histone and DNA Modification\\
\noalign{\smallskip}\hline\noalign{\smallskip}
\end{tabular}
$^a$ Table foot note (with superscript)
\end{table}
%

%
%

\Extrachap{Glossary}

Use the template \emph{glossary.tex} together with the Springer document class SVMono (monograph-type books) or SVMult (edited books) to style your glossary\index{glossary} in the Springer layout.

\runinhead{glossary term} Write here the description of the glossary term. Write here the description of the glossary term. Write here the description of the glossary term.

\runinhead{glossary term} Write here the description of the glossary term. Write here the description of the glossary term. Write here the description of the glossary term.

\runinhead{glossary term} Write here the description of the glossary term. Write here the description of the glossary term. Write here the description of the glossary term.

\runinhead{glossary term} Write here the description of the glossary term. Write here the description of the glossary term. Write here the description of the glossary term.

\runinhead{glossary term} Write here the description of the glossary term. Write here the description of the glossary term. Write here the description of the glossary term.
\printindex


\end{document}